\documentclass[11pt,a4paper]{article}
\usepackage[table,xcdraw]{xcolor}
\usepackage[hyperref]{acl2020}
\usepackage{nameref}
\usepackage{times}
\usepackage{latexsym}
\usepackage{graphics}
\usepackage{amsmath}
\usepackage{amsfonts}
\usepackage{algorithm2e}
\usepackage{commath}
\usepackage{bbm}
\usepackage{cleveref}
\usepackage{graphicx}
\usepackage{tabularx}
\usepackage{boxedminipage}
\newcolumntype{L}{>{\arraybackslash}m{12cm}}
\crefformat{section}{\S#2#1#3}
\crefformat{subsection}{\S#2#1#3}
\crefformat{subsubsection}{\S#2#1#3}

\usepackage{microtype}
\aclfinalcopy 

\title{Neural Data-to-Text Generation via Jointly Learning the Segmentation and Correspondence}

\author{Xiaoyu Shen$^{1,2}$\thanks{\hspace{1.5 mm}
Correspondence can be sent to 
xshen@mpi-inf.mpg.de}, Ernie Chang$^{3}$, Hui Su$^{4}$, Jie Zhou$^{4}$ and Dietrich Klakow$^{1}$\\
  $^{1}$Spoken Language Systems (LSV)\hspace{1.5 mm} $^{2}$Max Planck Institute for Informatics\\
  $^{3}$Department of Language Science and Technology, Saarland Informatics Campus\\
  $^{4}$Pattern Recognition Center, Wechat AI, Tencent Inc, China}

\date{}
\begin{document}
\maketitle
\begin{abstract}
The neural attention model has achieved great success in data-to-text generation tasks. 
Though usually excelling at producing fluent text, it suffers from the problem of information missing, repetition and ``hallucination''. 
Due to the black-box nature of the neural attention architecture, avoiding these problems in a systematic way is non-trivial. 
To address this concern, we propose to explicitly segment target text into fragment units and align them with their data correspondences. 
The segmentation and correspondence are jointly learned as latent variables without any human annotations. 
We further impose a soft statistical constraint to regularize the segmental granularity. 
The resulting architecture maintains the same expressive power as neural attention models, while being able to generate fully interpretable outputs with several times less computational cost. 
On both E2E and WebNLG benchmarks, we show the proposed model consistently outperforms its neural attention counterparts.
\end{abstract}

\section{Introduction}
Data-to-text generation aims at automatically producing natural language descriptions of structured database~\cite{reiter1997building}. 
Traditional statistical methods usually tackle this problem by breaking the generation process into a set of local decisions that are learned separately~\cite{belz2008automatic,angeli2010simple,kim2010generative,oya2014template}. 
Recently, neural attention models conflate all steps into a single end-to-end system and largely simplify the training process~\cite{mei2016talk,lebret2016neural,shen2017estimation,su2018nexus,su2019improving,chang2020unsupervised}. 
However, the black-box conflation also renders the generation uninterpretable and hard to control~\cite{wiseman2018learning,shen2019select}. 
Verifying the generation correctness in a principled way is non-trivial. 
In practice, it often suffers from the problem of information missing, repetition and ``hallucination''~\cite{duvsek2018findings,duvsek2020evaluating}.
\begin{figure}[h]
\centering
\fbox{
\parbox{0.9\linewidth}{
\textbf{Source data}:
\vspace{0.1cm}

Name[Clowns], PriceRange[more than \pounds 30],\\
EatType[pub], FamilyFriendly[no]

\vspace{0.2cm}

\textbf{Generation:}

\large \textcircled{\small 1}\underline{Name} $\,\to\,$ \large \textcircled{\small 2}(Clowns)

\large \textcircled{\small 3}\underline{FamilyFriendly} $\,\to\,$ \large \textcircled{\small 4}(is a child-free)

\large \textcircled{\small 5}\underline{PriceRange} $\,\to\,$ \large \textcircled{\small 6}(, expensive)

\large \textcircled{\small 7}\underline{EatType} $\,\to\,$ \large \textcircled{\small 8}(pub.)
}}
    \caption{Generation from our model on the E2E dataset. Decoding is performed segment-by-segment. Each segment realizes one data record. {\large{\textcircled{\small 1}}\raisebox{-0.9ex}{\~{}}\large{\textcircled{\small 8}}} mark the decision order in the generation process.
    }
    \label{fig: intro}
\end{figure}

In this work, we propose to explicitly exploit the \emph{segmental structure} of text. 
Specifically, we assume the target text is formed from a sequence of segments. 
Every segment is the result of a two-stage decision: 
(1) Select a proper data record to be described and 
(2) Generate corresponding text by \emph{paying attention only to the selected data record}. 
This decision is repeated until all desired records have been realized. 
Figure~\ref{fig: intro} illustrates this process.

Compared with neural attention, the proposed model has the following advantages: 
(1) We can monitor the corresponding data record for every segment to be generated. 
This allows us to easily control the output structure and verify its correctness\footnote{For example, we can perform a similar constrained decoding as in \citet{balakrishnan-etal-2019-constrained} to rule out outputs with undesired patterns.}. 
(2) Explicitly building the correspondence between segments and data records can potentially reduce the hallucination, as noted in \cite{wu2018hard,deng2018latent} that hard alignment usually outperforms soft attention. 
(3) When decoding each segment, the model pays attention only to the selected data record instead of averaging over the entire input data. 
This largely reduces the memory and computational costs~\footnote{Coarse-to-fine attention~\cite{ling2017coarse,deng2017image} was proposed for the same motivation, but they resort to reinforcement learning which is hard to train, and the performance is sacrificed for efficiency.}.

To train the model, we \emph{do not} rely on any human annotations for the segmentation and correspondence, but rather marginalize over all possibilities to maximize the likelihood of target text, which can be efficiently done within polynomial time by dynamic programming. 
This is essentially similar to traditional methods of inducing segmentation and alignment with semi-markov models~\cite{daume2005induction,liang2009learning}. 
However, they make strong independence assumptions thus perform poorly as a generative model~\cite{angeli2010simple}. In contrast, the transition and generation in our model condition on \emph{all previously generated text}. 
By integrating an autoregressive neural network structure, our model is able to capture unbounded dependencies while still permitting tractable inference. 
The training process is stable as it does not require any sampling-based approximations. 
We further add a soft statistical constraint to control the segmentation granularity via posterior regularization~\cite{ganchev2010posterior}. On both the E2E and WebNLG benchmarks, our model is able to produce significantly higher-quality outputs while being several times computationally cheaper. 
Due to its fully interpretable segmental structure, it can be easily reconciled with heuristic rules or hand-engineered constraints to control the outputs~\footnote{Code to be released soon}.

\section{Related Work}
\label{sec: related}
Data-to-text generation is traditionally dealt with using a pipeline structure containing content planning, sentence planning and linguistic realization~\cite{reiter1997building}. 
Each target text is split into meaningful fragments and aligned with corresponding data records, either by hand-engineered rules~\cite{kukich1983design,mckeown1992text} or statistical induction~\cite{liang2009learning,koncel2014multi,qin2018learning}. 
The segmentation and alignment are used as supervision signals to train the content and sentence planner~\cite{barzilay2005collective,angeli2010simple}. 
The linguistic realization is usually implemented by template mining from the training corpus~\cite{kondadadi2013statistical,oya2014template}. 
Our model adopts a similar pipeline generative process, but integrates all the sub-steps into a single end-to-end trainable neural architecture. 
It can be considered as a neural extension of the PCFG system in \citet{konstas2013global}, with a more powerful transition probability considering inter-segment dependence and a state-of-the-art attention-based language model as the linguistic realizer. 
\citet{wiseman2018learning} tried a similar neural generative model to induce templates. 
However, their model only captures loose data-text correspondence and adopts a weak markov assumption for the segment transition probability. 
Therefore, it underperforms the neural attention baseline as for generation. Our model is also in spirit related to recent attempts at separating content planning and surface realization in neural data-to-text models~\cite{zhao2018comprehensive,puduppully2019data,moryossef2019step, ferreira2019neural}. 
Nonetheless, all of them resort to \emph{manual annotations or hand-engineered rules applicable only for a narrow domain}. 
Our model, instead, automatically learn the optimal content planning via exploring over exponentially many segmentation/correspondence possibilities.

There have been quite a few neural alignment models applied to tasks like machine translation~\cite{wang-etal-2018-neural-hidden,deng2018latent}, character transduction~\cite{wu2018hard,shankar2018posterior} and summarization~\cite{yu2016online,shen2019improving}. 
Unlike word-to-word alignment, we focus on learning the alignment between data records and text segments.
Some works also integrate neural language models to jointly learn the segmentation and correspondence, e.g., phrase-based machine translation~\cite{huang2017towards}, speech recognition~\cite{wang2017sequence} and vision-grounded word segmentation~\cite{kawakami2018unsupervised}. 
Data-to-text naturally fits into this scenario since each data record is normally verbalized in one continuous text segment.

\section{Background: Data-to-Text}
Let $X,Y$ denote a source-target pair. $X$ is structured data containing a set of records and $Y$ corresponds to $y_1,y_2,\ldots,y_m$ which is a text description of $X$. The goal of data-to-text generation is to learn a distribution $p(Y|X)$ to automatically generate proper text describing the content of the data.

The neural attention architecture handles this task with an encode-attend-decode process~\cite{bahdanau2015neural}. 
The input $X$ is processed into a sequence of $x_1,x_2,\ldots,x_n$, normally by flattening the data records~\cite{wiseman2017challenges}. 
The encoder encodes each $x_i$ into a vector $h_i$. 
At each time step, the decoder attends to encoded vectors and outputs the probability of the next token by $p(y_t|y_{1:t-1},A_t)$. $A_t$ is a weighted average of source vectors:
\begin{equation}
\label{eq: attention}
\begin{split}
A_t &= \sum_{i}\alpha_{t,i}h_{i}\\
\alpha_{t,i} &= \frac{e^{f(h_{i}, d_t)}}{\sum_j e^{f(h_{j}, d_t)}}
\end{split}
\end{equation}
$d_{t}$ is the hidden state of the decoder at time step $t$. 
$f$ is a score function to compute the similarity between  $h_i$ and $d_t$~\cite{luong2015effective}.

\section{Approach}

\begin{figure}[t!]
  \centering
\includegraphics[width=0.9\columnwidth]{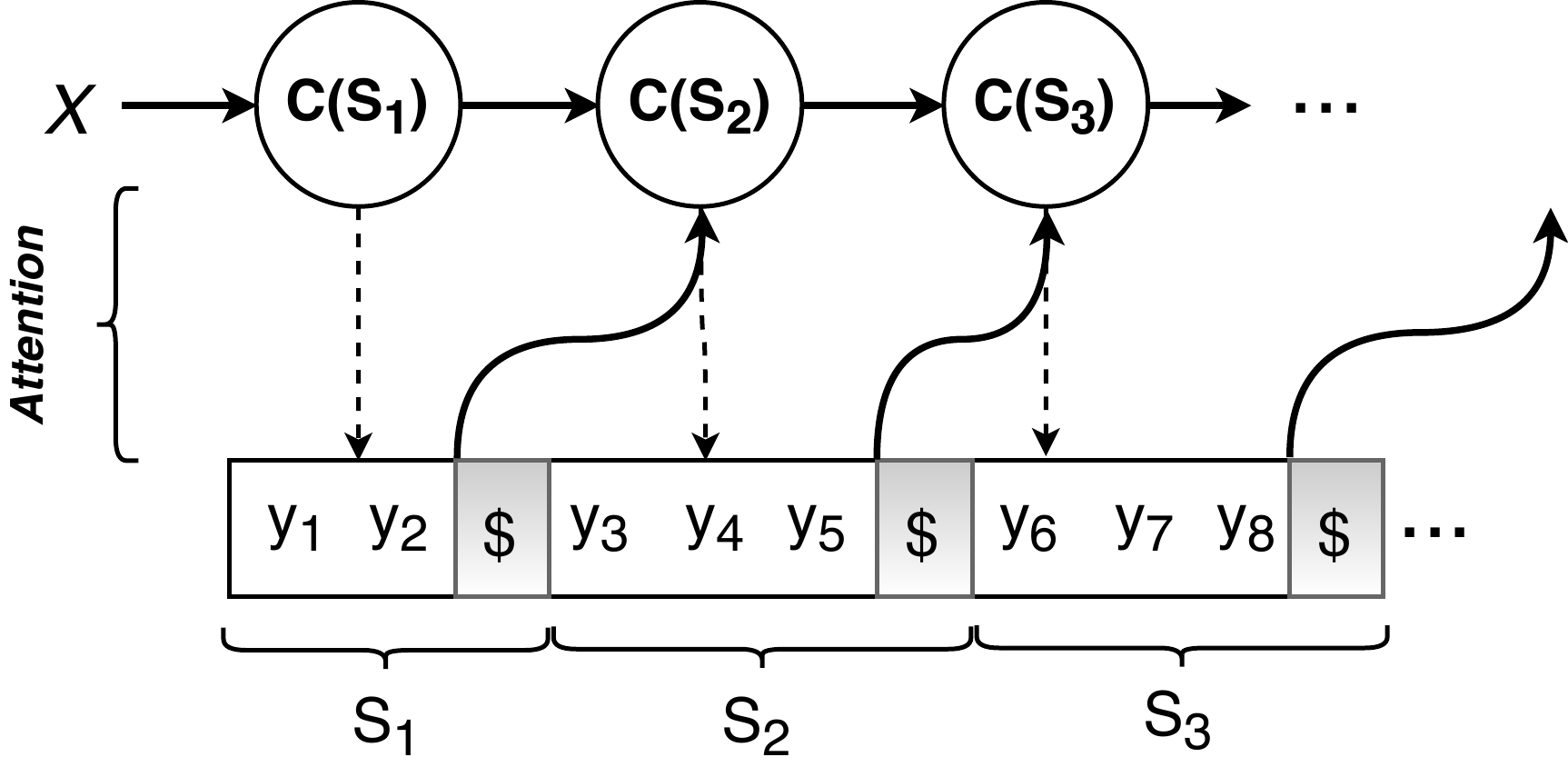}
\caption{\small{Generation process of our approach. 
Segment end symbol $\$$ is ignored when updating the state of the decoder. 
\textit{Solid arrows} indicate the transition model and \textit{dashed arrows} indicate the generation model. 
Every segment $s_o$ is generated by attending only to the corresponding data record $c(s_o)$.}}
\label{fig: architecture}
\end{figure}

Suppose the input data $X$ contains a set of records $r_1, r_2,..., r_K$. 
Our assumption is that the target text $y_{1:m}$ can be segmented into a sequence of fragments. 
Each fragment corresponds to one data record. As the ground-truth segmentation and correspondence are not available, we need to enumerate over all possibilities to compute the likelihood of $y_{1:m}$. 
Denote by $\mathcal{S}_y$ the set containing all valid segmentation of $y_{1:m}$. 
For any valid segmentation $s_{1:\tau_{s}}\in \mathcal{S}_y$, $\pi(s_{1:\tau_{s}})=y_{1:m}$, where $\pi$ means concatenation and $\tau_{s}$ is the number of segments. 
For example, let $m=5$ and $\tau_{s}=3$. 
One possible segmentation would be $s_{1:\tau_{s}}=\{\{ y_1, y_2, \$ \}, \{ y_3, \$ \}, \{ y_4, y_5, \$ \}\}$. $\$$ is the end-of-segment symbol and is removed when applying the $\pi$ operator. 
We further define $c(*)$ to be the corresponding data record(s) of $*$. 
The likelihood of each text is then computed by enumerating over all possibilities of $s_{1:\tau_{s}}$ and $c(s_{1:\tau_{s}})$:
\begin{equation}
\label{eq: marginal}
\begin{split}
&p(y_{1:m}|X) = \sum_{s_{1:\tau_{s}}\in \mathcal{S}_y}p(s_{1:\tau_{s}}|X)\\
 &=\sum_{s_{1:\tau_{s}}\in \mathcal{S}_y}\prod_{o=1}^{\tau_{s}}\sum_{c(s_{o})=r_1}^{r_K}p(s_{o}|\pi(s_{<o}),c(s_{o}))\\&\times p(c(s_o)|\pi(s_{<o}),c(s_{<o}))
\end{split}
\end{equation}
Every segment is generated by first selecting the data record based on the \emph{transition probability} $p(c(s_o)|\pi(s_{<o}),c(s_{<o}))$, then generating tokens based on the word \emph{generation probability} $p(s_{o}|\pi(s_{<o}),c(s_{o}))$. Figure~\ref{fig: architecture} illustrates the generation process of our model.

\paragraph{Generation Probability} We base the generation probability on the same decoder as in neural attention models. The only difference is that \emph{the model can only pay attention to its corresponding data record}. The attention scores of other records are masked out when decoding $s_o$:
\begin{equation*}
    \alpha_{t,i} = \frac{e^{f(h_{i}, d_t)}\mathbbm{1}(x_i\in c(s_o))}{\sum_j e^{f(h_{j}, d_t)}\mathbbm{1}(x_j\in c(s_o))}
\end{equation*}
where $\mathbbm{1}$ is the indicator function. This forces the model to learn proper correspondences and enhances the connection between each segment and the data record it describes.

Following the common practice, we define the output probability with the pointer generator~\cite{see2017get,wiseman2017challenges}:
\begin{equation*}
\begin{split}
p_{gen} &= \sigma(\text{MLP}_{g}([d_{t}\circ A_{t}]))\\
p_{vocab} &= \text{softmax}(W_1 d_t + W_2 A_t)\\
p_\theta(y_{t}|y_{<t}) &= p_{gen}p_{vocab}(y_t)+(1-p_{gen})\sum_{i:y_t=x_i}a_{t,i}
\end{split}
\end{equation*} 
$d_t$ is the decoder's hidden state at time step $t$. $\circ$ denotes vector concatenation. $A_t$ is the context vector. MLP indicates multi-layer perceptron and $\sigma$ normalizes the score between $(0,1)$. $W_1$ and $W_2$ are trainable matrices. $p_{gen}$ is the probability that the word is generated from a fixed vocabulary distribution $p_{vocab}$ instead of being copied. The final decoding probability $p_\theta(y_t)$ is marginalized over $p_{vocab}$ and the copy distribution. The generation probability of $s_o$ factorizes over the words within it and the end-of-segment token:
\begin{equation*}
    p(s_{o}|\pi(s_{<o}),c(s_{o})) =  p_\theta(\$|y_{1:t})\times \prod_{y_t \in s_o} p_\theta(y_t|y_{<t})
\end{equation*}

\paragraph{Transition Probability} We make a mild assumption that $c(s_o)$ is dependent only on $c(s_{o-1})$ and $\pi(s_{1:o-1})$ but irrelevant of $c(s_{<o-1})$, which is a common practice when modelling alignment~\cite{och1999improved,yu2016online,shankar2018posterior}. The transition probability is defined as:
\begin{equation}
\label{eq: transit}
\begin{split}
&p(c(s_o)=r_i|c(s_{<o}), \pi(s_{<o}))\\ \approx & p(c(s_o)=r_i|c(s_{o-1}), \pi(s_{<o}))\\
 \propto&f(r_i)^T [M^T A_{s_{o-1}} + N^T d_{s_{o-1}}]
\end{split}
\end{equation}
A softmax layer is finally applied to the above equation to normalize it as a proper probability distribution. $f(r_i)$ is a representation of $r_i$, which is defined as a max pooling over all the word embeddings contained in $r_i$. $A_{s_{o-1}}$ is the attention context vector when decoding the last token in $s_{o-1}$, defined as in Equation~\ref{eq: attention}. It carries important information from $c(s_{o-1})$ to help predict $c(s_o)$. $d_{s_{o-1}}$ is the hidden state of the neural decoder which goes through all history tokens $\pi(s_{1:o-1})$. $M,N$ are trainable matrices to project $A_{s_{o-1}}$ and $d_{s_{o-1}}$ into the same dimension as $f(r_i)$.

We further add one constraint to prohibit \emph{self-transition}, which can be easily done by zeroing out the transition probability in Equation~\ref{eq: transit} when $c(s_{o})=c(s_{o-1})$. This forces the model to group together text describing the same data record.

Since Equation~\ref{eq: transit} conditions on all previously generated text, it is able to capture more complex dependencies as in semi-markov models~\cite{liang2009learning,wiseman2018learning}.

\paragraph{Null Record} In our task, we find some frequent phrases, e.g., ``it is", ``and", tend to be wrongly aligned with some random records, similar to the garbage collection issue in statistical alignment~\cite{brown1993mathematics}. This hurt the model interpretability. Therefore, we introduce an additional null record $r_0$ to attract these non-content phrases. The context vector when aligned to $r_0$ is a zero vector so that the decoder will decode words based solely on the language model without relying on the input data.
\paragraph{Training} Equation~\ref{eq: marginal} contains exponentially many combinations to enumerate over. Here we show how to efficiently compute the likelihood with the forward algorithm in dynamic programming~\cite{rabiner1989tutorial}. We define the forward variable $\alpha(i,j)=p(y_{1:i},c(y_i)=j|X)$. With the base $\alpha(1,j)=p(y_1|c(y_{1})=j)$. The recursion goes as follows for $i=1,2,\ldots,m-1$:
\begin{equation}
\label{eq: forward}
\begin{split}
\alpha&(i+1,j)=\sum_{p=1}^{i}\sum_{q=r_0}^{r_K}\alpha(p,q)\\&\times p(c(y_{p+1})=j|c(y_p)=q, y_{1:p})\\
&\times p(y_{p+1:i+1}|c(y_{p+1:i+1})=q, y_{1:p})\\
&\times p(\$|c(y_{p+1:i+1})=q, y_{1:i+1})
\end{split}
\end{equation}
The final likelihood of the target text can be computed as $p(y_{1:m}|X)=\sum_{j=r_0}^{r_K}\alpha(m,j)$. As the forward algorithm is fully differentiable, we maximize the log-likelihood of the target text by backpropagating through the dynamic programming. The process is essentially equivalent to the generalized EM algorithm~\cite{eisner2016inside}. By means of the modern automatic differentiation tools, we avoid the necessity to calculate the posterior distribution manually~\cite{kim2018tutorial}.

To speed up training, we set a threshold $L$ to the maximum length of a segment as in \citet{liang2009learning,wiseman2018learning}. This changes the complexity in Equation~\ref{eq: forward} to a constant $O(LK)$ instead of scaling linearly with the length of the target text. Moreover, as pointed out in \citet{wang2017sequence}, the computation for the longest segment can be reused for shorter segments. We therefore first compute the generation and transition probability for the whole sequence in one pass. The intermediate results are then cached to efficiently proceed the forward algorithm without any re-computation.

One last issue is the numerical precision, it is important to use the log-space binary operations to avoid underflow~\cite{kim2017structured}.

\begin{table}[ht]
\small
\centering
\begin{tabular}{@{}l@{}}
Near[riverside], Food[French], EatType[pub], Name[Cotto]\\
\hline
\begin{tabular}[c]{@{}l@{}}
1. [Near]$_{\text{Near}}$[the]$_{\text{Null}}$[riverside]$_{\text{Near}}$[is a]$_{\text{Null}}$[French]$_{\text{Food}}$\\
\; [pub]$_{\text{EatType}}$[called]$_{\text{Null}}$[Cotto]$_{\text{Name}}$[.]$_{\text{Null}}$ \\
2. [Near the riverside]$_{\text{Near}}$[is]$_{\text{Null}}$[a French]$_{\text{Food}}$[pub]$_{\text{EatType}}$\\
\; [called Cotto]$_{\text{Name}}$[.]$_{\text{Null}}$ \\
3. [Near the riverside]$_{\text{Near}}$[is a French]$_{\text{Food}}$[pub]$_{\text{EatType}}$\\
\; [called Cotto .]$_{\text{Name}}$ \\
4. [Near the riverside]$_{\text{Near}}$[is a French pub]$_{\text{Food}}$\\
\; [called Cotto .]$_{\text{Name}}$
\end{tabular} \\
\hline
\end{tabular}
\caption{\small Segmentation with various granularities. 1 is too fine-grained while 4 is too coarse. We expect a segmentation like 2 or 3 to better control the generation.}
\label{tab: granularity}
\end{table}

\paragraph{Segmentation Granularity}
\label{para: granu}
There are several valid segmentations for a given text. As shown in Table~\ref{tab: granularity}, when the segmentation (example 1) is too fine-grained, controlling the output information becomes difficult because the content of one data record is realized in separate pieces~\footnote{The finer-grained segmentation might be useful if the focus is on modeling the detailed discourse structure instead of the information accuracy~\cite{reed2018can,balakrishnan-etal-2019-constrained}, which we leave for future work.}. When it is too coarse, the alignment might become less accurate (as in Example 4, ``pub" is wrongly merged with previous words and aligned together to the ``Food" record). In practice, we expect the segmentation to stay with accurate alignment yet avoid being too brokenly separated. To control the granularity as we want, we utilize posterior regularization~\cite{ganchev2010posterior} to constrain the expected number of segments for each text~\footnote{We can also utilize some heuristic rules to help segmentation. For example, we can prevent breaking syntactic elements obtained from an external parser~\cite{yang-etal-2019-low} or match entity names with handcrafted rules~\cite{chen2018sheffield}. The interpretability of the segmental structure allows easy combination with these rules. We focus on a general \emph{domain-agnostic} method in this paper, though heuristic rules might bring further improvement under certain cases.}, which can be calculated by going through a similar forward pass as in Equation~\ref{eq: forward}~\cite{eisner2002parameter,backes2018simulating}. Most computation is shared without significant extra burden. The final loss function is:
\begin{equation}
\label{eq: loss}
-\log \mathbbm{E}_{\mathcal{S}_y}p(s_{1:\tau_s}|X) + \max(\abs{\mathbbm{E}_{\mathcal{S}_y}\tau_s-\eta}, \gamma)
\end{equation}
$\log \mathbbm{E}_{\mathcal{S}_y}p(s_{1:\tau_s}|X)$ is the log-likelihood of target text after marginalizing over all valid segmentations. $\mathbbm{E}_{\mathcal{S}_y}\tau_s$ is the expected number of segments and $\eta, \gamma$ are hyperparameters. We use the max-margin loss to encourage $\mathbbm{E}_{\mathcal{S}_y}\tau_s$ to stay close to $\eta$ under a tolerance range of $\gamma$. 


\paragraph{Decoding}
\label{para: decoding}
The segment-by-segment generation process allows us to easily constrain the output structure. Undesirable patterns can be rejected before the whole text is generated. We adopt three simple constraints for the decoder:
\begin{enumerate}
    \item Segments must not be empty.
    \item The same data record cannot be realized more than once (except for the null record).
    \item The generation will not finish until all data records have been realized.
\end{enumerate}
Constraint 2 and 3 directly address the information repetition and missing problem. When segments are incrementally generated, the constraints will be checked against for validity. Note that adding the constraints hardly incur any cost, the decoding process is still finished \emph{in one pass}. No post-processing or reranking is needed.

\paragraph{Computational Complexity} Suppose the input data has $M$ records and each record contains $N$ tokens. The computational complexity for neural attention models is $O(MN)$ at each decoding step where the whole input is retrieved. Our model, similar to chunkwise attention~\cite{chiu2017monotonic} or coarse-to-fine attention~\cite{ling2017coarse}, reduces the cost to $O(M+N)$, where we select the record in $O(M)$ at the beginning of each segment and attend only to the selected record in $O(N)$ when decoding every word. For larger input data, our model can be significantly cheaper than neural attention models.
\section{Experiment Setup}
\paragraph{Dataset}
We conduct experiments on the E2E~\cite{novikova2017e2e} and WebNLG~\cite{colin-etal-2016-webnlg} datasets. E2E is a crowd-sourced dataset containing 50k instances in the restaurant domain. The inputs are dialogue acts consisting of three to eight slot-value pairs. WebNLG contains 25k instances describing entities belonging to fifteen distinct DBpedia categories. The inputs are up to seven RDF triples of the form \emph{(subject, relation, object)}. 
\paragraph{Implementation Details}
We use a bi-directional LSTM encoder and uni-directional LSTM decoder for all experiments. Input data records are concatenated into a sequence and fed into the encoder. We choose the hidden size of encoder/decoder as 512 for E2E and 256 for WebNLG. The word embedding is with size 100 for both datasets and initialized with the pre-trained Glove embedding~\footnote{\url{nlp.stanford.edu/data/glove.6B.zip}}~\cite{pennington2014glove}. We use a drop out rate of $0.3$ for both the encoder and decoder. Models are trained using the Adam optimizer~\cite{kingma2014adam} with batch size 64. The learning rate is initialized to 0.01 and decays an order of magnitude once the validation loss increases. All hyperparameters are chosen with grid search according to the validation loss. Models are implemented based on the open-source library PyTorch~\cite{pytorch}.
We set 
the hyperparameters in Eq.~\ref{eq: loss} as $\eta=K, \gamma=1$ (recall that $K$ is the number of records in the input data). The intuition is that every text is expected to realize the content of all $K$ input records. It is natural to assume every text can be roughly segmented into $K$ fragments, each corresponding to one data record. A deviation of $K\pm 1$ is allowed for noisy data or text with complex structures.
\paragraph{Metrics}
We measure the quality of system outputs from three perspectives: (1) \emph{word-level overlap} with human references, which is a commonly used metric for text generation. We report the scores of BLEU-4~\cite{papineni2002bleu}, ROUGE-L~\cite{lin2004rouge}, Meteor~\cite{banerjee2005meteor} and CIDEr~\cite{vedantam2015cider}
. (2) \emph{human evaluation}. Word-level overlapping scores usually correlate rather poorly with human judgements on fluency and information accuracy~\cite{reiter2009investigation,novikova2017we}. Therefore, we passed the input data and generated text to human annotators to judge if the text is fluent by grammar (scale 1-5 as in \citet{belz2006comparing}), contains wrong fact inconsistent with input data, repeats or misses information. We report the \emph{averaged score} for fluency and \emph{definite numbers} for others. The human is conducted on a sampled subset from the test data. To ensure the subset covers inputs with all possible number of records ($K\in[3,8]$ for E2E and $K\in[1,7]$ for WebNLG), we sample 20 instances for every possible $K$. Finally,we obtain 120 test cases for E2E and 140 for WebNLG~\footnote{The original human evaluation subset of WebNLG is randomly sampled, most of the inputs contain less than 3 records, so we opt for a new sample for a thorough evaluation.}. (3) \emph{Diversity of outputs}. Diversity is an important concern for many real-life applications. We measure it by the number of unique unigrams and trigrams over system outputs, as done in \citet{duvsek2020evaluating}.
\section{Results}
\begin{figure}[t]
  \centering
\includegraphics[width=\columnwidth]{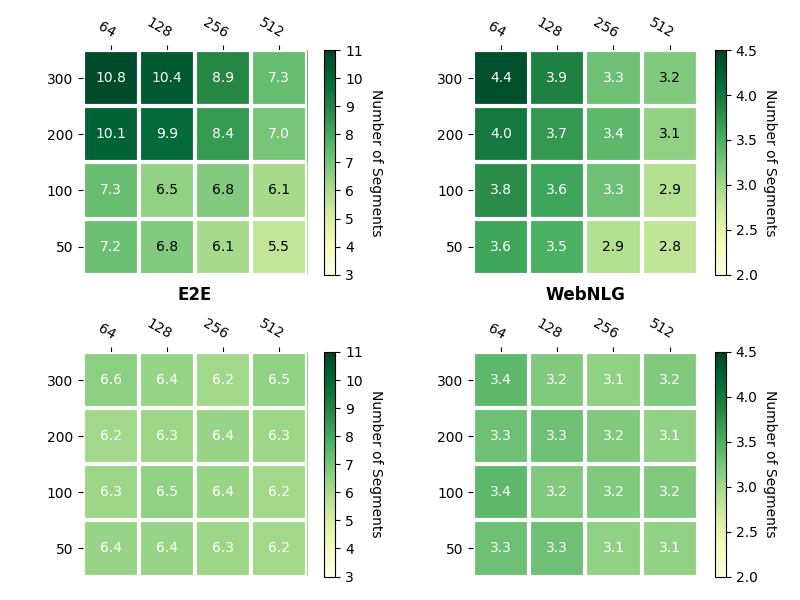}
\caption{\small Average expected number of segments with varying hyperparameters. x-axis is the encoder/decoder hidden size and y-axis is the word embedding size. Upper two figures are without the granularity regularization and the bottom two are with regularization.}
\label{fig: n_segment}
\end{figure}
\begin{table*}[t!]
  \small
  \begin{center}
    \begin{tabularx}{0.93\linewidth}{l|cccc|cccc|cc}
    \hline 
     \textbf{Metrics} &
     \multicolumn{4}{c|}{\textbf{Word Overlap}} & \multicolumn{4}{c|}{\textbf{Human Evaluation}} & \multicolumn{2}{c}{\textbf{Diversity}} \\ 
     \textbf{Models} & \textbf{BLEU} & \textbf{R-L} & \textbf{Meteor} & \textbf{CIDEr} & \textbf{Fluent} & \textbf{Wrong} & \textbf{Repeat} & \textbf{Miss}  & \textbf{Dist-1} & \textbf{Dist-3} \\
     \hline
     \textbf{\textsc{Slug}}  & \textbf{0.662} & \textbf{0.677} & 0.445 & \textbf{2.262} & 4.94 & 5 & \textbf{0} & 17 & 74 & 507 \\
     \textbf{\textsc{DANGNT}} & 0.599 & 0.663  & 0.435 & 2.078 & 4.97 & \textbf{0} & \textbf{0} & 21 & 61 & 301 \\
     \textbf{\textsc{TUDA}}  & 0.566 & 0.661  & \textbf{0.453} & 1.821 & \textbf{4.98} & \textbf{0} & \textbf{0} & \textbf{10} & 57  & 143 \\
      \textbf{\textsc{N\_Temp}}  & 0.598 & 0.650  & 0.388 & 1.950 & 4.84 & 19 & 3 & 35 & \textbf{119} & \textbf{795}\\
     \hline
     \textbf{\textsc{PG}} & 0.638 & 0.677 & 0.449 & 2.123 & 4.91 & 15 &  1 & 29 & 133 & 822 \\
    \textbf{\textsc{Ours}}  & 0.647 & \textbf{0.683}  & 0.453 & 2.222 & \textbf{4.96} & \textbf{0} & 1 & 15 & 127 & 870\\
     \textbf{\textsc{Ours (+R)}}  & 0.645 & 0.681  & 0.452 & 2.218 & 4.95 & \textbf{0} & \textbf{0} & 13 & 133  & 881\\
     \textbf{\textsc{Ours (+RM)}}  & \textbf{0.651} & 0.682 & \textbf{0.455} & \textbf{2.241} & 4.95 & \textbf{0} & \textbf{0} & \textbf{3} & \textbf{135} & \textbf{911}\\
     \hline
    \end{tabularx}
  \end{center}
   \caption{\small Automatic and human evaluation results on E2E dataset. \textbf{SLUG}, \textbf{DANGNT}, \textbf{TUDA} and \textbf{N\_TEMP} are from previous works and the other models are our own implementations.}
     \label{tab: e2eresults}%
     \vspace{-0.3cm}
\end{table*}%
\begin{table*}[t!]
  \small
  \begin{center}
    \begin{tabularx}{0.95\linewidth}{l|cccc|cccc|cc}
    \hline 
     \textbf{Metrics} &
     \multicolumn{4}{c|}{\textbf{Word Overlap}} & \multicolumn{4}{c|}{\textbf{Human Evaluation}} & \multicolumn{2}{c}{\textbf{Diversity}} \\ 
     \textbf{Models} & \textbf{BLEU} & \textbf{R-L} & \textbf{Meteor} & \textbf{CIDEr} & \textbf{Fluent} & \textbf{Wrong} & \textbf{Repeat} & \textbf{Miss}  & \textbf{Dist-1} & \textbf{Dist-3} \\
     \hline
     \textbf{\textsc{Melbourne}}  & 0.450 & \textbf{0.635} & 0.376 & \textbf{2.814} & 4.16 & 42 & 22 & 37 & 3167 & \textbf{13,744} \\
     \textbf{\textsc{UPF-FORGe}} & 0.385 & 0.609  & 0.390 & 2.500 & 4.08 & \textbf{29} & \textbf{6} & \textbf{28} & \textbf{3191} & 12,509 \\\hline
     \textbf{\textsc{PG}} & 0.452 & 0.652 & 0.384 & 2.623 & 4.13 & 43 &  26 & 42 & 3,218 & 13,403 \\
    \textbf{\textsc{Ours}}  & 0.453 & 0.656  & 0.388 & 2.610 & 4.23 & 26 & 19 & 31 & 3,377  & 14,516\\
     \textbf{\textsc{Ours (+R)}}  & 0.456 & \textbf{0.657}  & 0.390 & \textbf{2.678} & \textbf{4.28} & \textbf{18} & \textbf{2} & 24 & 3,405 & 14,351\\
     \textbf{\textsc{Ours (+RM)}}  & \textbf{0.461} & 0.654  & \textbf{0.398} & 2.639 & 4.26 & 23 & 4 & \textbf{5} & \textbf{3,457}  & \textbf{14,981}\\
     \hline
    \end{tabularx}
  \end{center}
   \caption{\small Automatic and human evaluation results on WebNLG dataset. \textbf{MELBOURNE} and \textbf{UPFUPF-FORGE} are from previous works and the other models are our own implementations.}
     \label{tab: webnlgresults}%
     \vspace{-0.3cm}
\end{table*}%
\begin{figure*}[ht]
\footnotesize{
\begin{center}
\resizebox{\textwidth}{!}{
\begin{tabular}{ll}
\cellcolor{red!30}{Egg Harbor Township, New Jersey \underline{isPartOf} New Jersey}&\cellcolor{orange!30}{Atlantic City International Airport \underline{Location Identifier} ``KACY" ICAO}\\\cellcolor{yellow!50}{Atlantic City International Airport \underline{location} Egg Harbor Township, New Jersey}&\cellcolor{green!40}{Egg Harbor Township, New Jersey \underline{country} United States}
 \\\cellcolor{blue!20}{Egg Harbor Township, New Jersey \underline{isPartOf} Atlantic County, New Jersey}&
 \\
\\
\end{tabular}
}
{\def\arraystretch{2}\tabcolsep=3pt
\begin{tabular}{l L}
PG         & Atlantic City International Airport is located in Egg Harbor Township , New Jersey , United States . \textbf{It is located in Egg Harbor Township , New Jersey} . \\ \hline
Ours & \colorbox{orange!30}{KACY is the ICAO location identifier of Atlantic City International Airport ,} \colorbox{yellow!50}{which is located at Egg Harbor Township , New jersey}\colorbox{green!40}{, in the United States]} \colorbox{orange!30}{\textbf{. The ICAO location identifier of Atlantic City International Airport is KACY .}} \\
\hline 
Ours (+R)     & \colorbox{orange!30}{KACY is the ICAO location identifier of Atlantic City International Airport ,} \colorbox{yellow!50}{which is located at Egg Harbor Township , New jersey}\colorbox{green!40}{, in the United States]} . \\
\hline
Ours (+RM) &  \colorbox{orange!30}{KACY is the ICAO location identifier of Atlantic City International Airport ,}\colorbox{yellow!50}{which is located at} \colorbox{yellow!50}{Egg Harbor Township , New jersey}\colorbox{green!40}{, in the United States} .\colorbox{blue!30}{The Egg Harbor Township is a part} \colorbox{blue!20}{of Atlantic County , New Jersey} . \colorbox{red!30}{Egg Harbor Township is a part of New Jersey .}
\end{tabular}}
\end{center}
\caption{\small Example generations from WebNLG. Relation types are \underline{underlined} and repeated generations are \textbf{bolded}. Segments and corresponding records in our model are marked in the same color. By adding explicit constraints to the decoding process, repetition and missing issues can be largely reduced. (better viewed in color)}
\label{fig: webnlgexample}
}
\end{figure*}
\begin{table*}[!htbp]
\fontsize{200pt}{300pt}\selectfont
\resizebox{\textwidth}{!}{
\begin{tabular}{|ll|}
\hline
Input:   & \colorbox{blue!5}{\colorbox{blue!15}{\strut [name the mill]}\colorbox{blue!18}{\strut [eattype} \colorbox{blue!30}{\strut restaurant]}\colorbox{blue!17}{\strut [food} \colorbox{blue!40}{\strut english]}\colorbox{blue!10}{\strut [pricerange} \colorbox{blue!7}{\strut moderate]}\colorbox{blue!8}{\strut [customerrating} \colorbox{blue!60}{\strut 1} \colorbox{blue!16}{\strut out of 5]}[area \colorbox{blue!17}{\strut riverside]}} ...  \\
PG: & the mill is a \textbf{\colorbox{blue!50}{\strut low} - priced} restaurant \textbf{in the city centre that delivers take - away} . it is located near caf\'e rouge.\\
Input:   & \colorbox{red!0}{\colorbox{blue!0}{[name the mill]}\colorbox{blue!0}{[eattype} \colorbox{blue!0}{restaurant]}\colorbox{blue!0}{[food} \colorbox{blue!0}{english]}\colorbox{red!10}{\strut [pricerange} \colorbox{red!50}{\strut moderate]}\colorbox{blue!0}{[customerrating} \colorbox{blue!0}{1} {out of 5]}\colorbox{blue!0}{[area}\colorbox{blue!0}{riverside]} ...}  \\ 
Ours:     &      [the mill][restaurant][near caf\'e rouge][in riverside][serves english food][at \colorbox{red!50}{\strut moderate} prices][. it is kid friendly and]...\\\hline      
\end{tabular}
}
\caption{\label{tab: attention}\small{(E2E) Attention map when decoding the word ``low" in the PG model and ``moderate" in our model. Hallucinated contents are \textbf{bolded}. The PG model wrongly attended to other slots thereby ``hallucinated" the content of ``low-priced". Our model always attends to one single slot instead of averaging over the whole inputs, the chance of hallucination is largely reduced.}}
\end{table*}
In this section, we first show the effects of the granularity regularization we proposed, then compare model performance on two datasets and analyze the performance difference. Our model is compared against the neural attention-based pointer generator (\textbf{PG}) which does not explicit learn the segmentation and correspondence. To show the effects of the constrained decoding mentioned in \cref{para: decoding}, \nameref{para: decoding}. we run our model with only the first constraint to prevent empty segments (denoted by \textbf{ours} in experiments), with the first two constraints to prevent repetition (denoted by \textbf{ours (+R)}), and with all constraints to further reduce information missing (denoted by \textbf{ours (+RM)}).
\paragraph{Segmentation Granularity}
We show the effects of the granularity regularization (\cref{para: granu}, \nameref{para: granu}) in Fig~\ref{fig: n_segment}. When varying the model size, the segmentation granularity changes much if no regularization is imposed. Intuitively if the generation module is strong enough (larger hidden size), it can accurately estimate the sentence likelihood itself without paying extra cost of switching between segments, then it tends to reduce the number of transitions. Vice versa, the number of transitions will grow if the transition module is stronger (larger embedding size). With the regularization we proposed, the granularity remains what we want regardless of the hyperparameters. We can thereby freely decide the model capacity without worrying about the difference of segmentation behavior.
\paragraph{Results on E2E}
 On the E2E dataset, apart from our implementations, we also compare agianst outputs from the \textbf{SLUG}~\cite{juraska2018deep}, the overall winner of the E2E challenge (seq2seq-based), \textbf{DANGNT}~\cite{nguyen2018structurebased}, the best grammar rule based model, \textbf{TUDA}~\cite{puzikov2018e2e}, the best template based model, and the autoregressive neural template model (\textbf{N\_TEMP}) from \citet{wiseman2018learning}. SLUG uses a heuristic slot aligner based on a set of handcrafted rules and combine a complex pipeline of data augmentation, selection, model ensemble and reranker, while our model has a simple end-to-end learning paradigm with no special delexicalizing, training or decoding tricks. 
 Table~\ref{tab: e2eresults} reports the evaluated results. Seq2seq-based models are more diverse than rule-based models at the cost of higher chances of making errors. As rule-based systems are by design always faithful to the input information, they made zero wrong facts in their outputs. Most models do not have the fact repetition issue because of the relatively simple patterns in the E2E dataset. therefore, adding the (+R) constraint only improves the performance minorly. The (+RM) constraint reduces the number of information missing to 3 without hurting the fluency. All the 3 missing cases are because of the wrong alignment between the period and one data record, which can be easily fixed by defining a simple rule. We put the error analysis in \cref{sec: error}.  N\_Temp performs worst among all seq2seq-based systems because of the restrictions we mentioned in \cref{sec: related}. As also noted by the author, it trades-off the generation quality for interpretability and controllability. In contrast, our model, despite relying on no heuristics or complex pipelines, \emph{made zero wrong facts with the lowest information missing rate, even surpassing rule-based models}. It also maintains interpretable and controllable without sacrificing the generation quality.
\paragraph{Results on WebNLG}
Table~\ref{tab: webnlgresults} reports the results evaluated on the WebNLG dataset. We also include results from \textbf{MELBOURNE}, a seq2seq-based system achieving highest scores on automatic metrics in the WebNLG challenge and \textbf{UPF-FORGE}, a classic grammar-based system that wins in the human evaluation
WebNLG contains significantly more distinct types of attributes than E2E, so the chance of making errors or repetitions increases greatly. Nevertheless, our model still \emph{performs on-par on automatic metrics with superior information adequacy and output diversity}. The (+R) decoding constraint becomes important since the outputs in WebNLG are much longer than those in E2E, neural network models have problems tracking the history generation beyond certain range. Models might repeat facts that have been already generated long back before. The (+R) constraint effectively reduces the repetition cases from 19 to 2. These 2 cases are intra-segment repetitions and failed to be detected since our model can only track inter-segment constraints (examples are in \cref{sec: error}). The (+RM) constraint brings down the information missing cases to 5 with slightly more wrong and repeated facts compared with (+R). Forcing models to keep generating until coveraging all records will inevitably increase the risk of making errors.
\paragraph{Discussions}
In summary, our models generates \emph{most diverse outputs, achieves similar or better performances in word-overlap automatic metrics while significantly reduces the information hallucination, repetition and missing problems}. An example of hallucination is shown in Table~\ref{tab: attention}. The standard PG model ``hallucinated" the contents of ``low-priced", ``in the city center" and ``delivers take-away". The visualized attention maps reveal that it failed to attend properly when decoding the word ``low". The decoding is driven mostly by language models instead of the contents of input data. In contrast, as we explicitly align each segment to one slot, the attention distribution of our model is \emph{concentrated on one single slot rather than averaged over the whole input}, the chance of hallucinating is therefore largely reduced.

Figure~\ref{fig: webnlgexample} shows some example generations from WebNLG. Without adding the decoding constraints, PG and our model both suffer from the problem of information repetition and missing. However, the interpretability of our model enables us to easily avoid these issues by constraining the segment transition behavior. For the attention-based PG model, there exists no simple way of applying these constraints. We can also explicitly control the output structure similar to \citet{wiseman2018learning}, examples are shown in \cref{sec: cont}.
\section{Conclusion}
In this work, we exploit the segmental structure in data-to-text generation. The proposed model significantly alleviates the information hallucination, repetition and missing problems without sacrificing the fluency and diversity. It is end-to-end trainable, domain-independent and allows explicit control over the structure of generated text. As our model is interpretable in the correspondence between segments and input records, it can be easily combined with hand-engineered heuristics or user-specific requirements to further improve the performance.
\section{Acknowledgements}
This research was funded in part by the DFG collaborative research center SFB 1102. Ernie Chang is supported by SFB 248 “Foundations of Perspicuous Software Systems” (E2); Xiaoyu Shen is supported by IMPRS-CS fellowship.
We sincerely thank the anonymous reviewers for their insightful comments that helped us to improve this paper.
\bibliography{anthology,acl2020}
\bibliographystyle{acl_natbib}

\clearpage

\appendix

\onecolumn

\section{Error Analysis}
\label{sec: error}
We analyze common errors below. 

\textbf{Missing:} Even with the coverage decoding constraint, the model can still occasionally miss information. 
We show one example in Table \ref{tab: miss}. The segments cover all input records, but the segment aligned to ``familyfriendly" only generates a period symbol. This happens 3 times on E2E and twice on WebNLG. On the other 3 cases of missing on WebNLG, some segments only generate one end-of-sentence symbol. Both conditions can be easily fixed by some simple filtering rules.

\textbf{Repeating:} There are still some repeating cases on the WebNLG dataset. Table~\ref{tab: repeat} shows one example. ``amsterdam-centrum is part of amsterdam" is repeated twice within a segment. As our constraint decoding can only prevent inter-segment repetition, it cannot fully avoid the repetition problem resulting from the intra-segment errors of RNNs.
\begin{table*}[!htbp]
\fontsize{200pt}{300pt}\selectfont
\resizebox{\textwidth}{!}{
\begin{tabular}{ll}
\hline
Input:   & \colorbox{red!20}{\strut name the phoenix} \colorbox{orange!30}{\strut eattype pub} \colorbox{yellow!60}{\strut food french} \colorbox{green!50}{\strut pricerange \pounds 20 - 25} \colorbox{gray!50}{\strut customerrating high} \colorbox{blue!20}{\strut area riverside}\\& \colorbox{red!60}{\strut familyfriendly yes} \colorbox{blue!40}{\strut near crowne plaza hotel}
  \\\hline
Output: & \colorbox{red!20}{\strut the phoenix} \colorbox{orange!30}{ \strut pub} \colorbox{blue!20}{\strut is located in riverside} \colorbox{blue!40}{\strut near crowne plaza hotel .} \colorbox{yellow!60}{\strut it serves french food} \colorbox{green!50}{\strut in the \pounds 20 -}\\& \colorbox{green!50}{\strut 25 price range} \colorbox{gray!50}{\strut . it has a high customer rating} \colorbox{red!60}{\strut .}\\
\hline      
\end{tabular}
}
\caption{\label{tab: miss}\small{Example of missing in E2E. The ``familyfriendly" is wligned to the period symbol.}}
\end{table*}

\begin{table*}[!htbp]
\fontsize{200pt}{300pt}\selectfont
\resizebox{\textwidth}{!}{
\begin{tabular}{|ll|}
\hline
Input:   & \colorbox{red!50}{\strut Amsterdam ground AFC Ajax (amateurs)} \colorbox{blue!40}{\strut Eberhard van der Laan leader Amsterdam} \colorbox{orange!30}{\strut Amsterdam-Centrum part Amsterdam}
  \\
Output: &  \colorbox{orange!30}{\strut amsterdam-centrum is part of amsterdam and amsterdam-centrum is part of amsterdam , the country where} \colorbox{blue!40}{\strut eberhard van}\\& \colorbox{blue!40}{\strut der laan is the leader and } \colorbox{red!50}{\strut the ground of afc ajax ( amateurs ) is located.} \\
\hline      
\end{tabular}
}
\caption{\label{tab: repeat}\small{(E2E) Example of repeatition in WebNLG. The phrase ``amsterdam-centrum is part of amsterdam" is repeated twice.}}
\end{table*}
    
\section{Controlling output structure}
\label{sec: cont}
As our model learns interpretable correspondence of each segment, it can control the output structures same as in \citet{wiseman2018learning}. Table~\ref{tab: control} shows example generations by sampling diverse segment structures.
\begin{table*}[!htbp]
\fontsize{200pt}{300pt}\selectfont
\resizebox{\textwidth}{!}{
\begin{tabular}{ll}
\hline
Input:   & \colorbox{red!20}{\strut name the phoenix} \colorbox{orange!30}{\strut eattype pub} \colorbox{yellow!60}{\strut food french} \colorbox{green!50}{\strut pricerange \pounds 20 - 25} \colorbox{gray!50}{\strut customerrating high} \colorbox{blue!20}{\strut area riverside}\\& \colorbox{red!60}{\strut familyfriendly yes} \colorbox{blue!40}{\strut near crowne plaza hotel}
  \\\hline
Output1: & \colorbox{red!20}{\strut the phoenix} \colorbox{orange!30}{ \strut pub} \colorbox{blue!20}{\strut is located in riverside} \colorbox{blue!40}{\strut near crowne plaza hotel .} \colorbox{yellow!60}{\strut it serves french food} \colorbox{green!50}{\strut in the \pounds 20 -}\\& \colorbox{green!50}{\strut 25 price range} \colorbox{gray!50}{\strut . it has a high customer rating} \colorbox{red!60}{\strut .}\\
Output2: & \colorbox{red!20}{\strut the phoenix} \colorbox{blue!20}{\strut is located in riverside} \colorbox{blue!40}{\strut near crowne plaza hotel .} \colorbox{red!60}{\strut it is a family - friendly} \colorbox{yellow!60}{\strut french} \colorbox{orange!30}{ \strut pub} \colorbox{green!50}{\strut with}\\& \colorbox{green!50}{\strut the price range of \pounds 20 - 25} \colorbox{gray!50}{\strut . it has a high customer rating .}\\
Output3: &  \colorbox{blue!20}{\strut located in riverside} \colorbox{blue!40}{\strut near crowne plaza hotel ,} \colorbox{red!20}{\strut the phoenix} \colorbox{yellow!60}{\strut is a french} \colorbox{orange!30}{ \strut pub} \colorbox{gray!50}{\strut with a high customer rating}\\& \colorbox{green!50}{\strut and a price range of \pounds 20 - 25.} \colorbox{red!60}{\strut It is family - friendly .}\\
\hline      
\end{tabular}
}
\caption{\label{tab: control}\small{Example of generations with diverse structures.}}
\end{table*}
\end{document}